# Implementation of Real-time Lane Detection on Autonomous Mobile Robot

Midriem Mirdanies
[a] *Telecommunication Software and Systems Research Group,*
*Faculty of Electrical Engineering,*
*Universiti Teknologi Malaysia (UTM),*
Johor Bahru, Johor, Malaysia
[b] *Research Center for Smart Mechatronics,*
*National Research and Innovation Agency (BRIN),*
Bandung, Indonesia
midriem@graduate.utm.my/ midr001@brin.go.id

Roni Permana Saputra
*Research Center for Smart Mechatronics,*
*National Research and Innovation Agency (BRIN),*
Bandung, Indonesia
roni008@brin.go.id

Edwar Yazid
*Research Center for Smart Mechatronics,*
*National Research and Innovation Agency (BRIN),*
Bandung, Indonesia
edwa005@brin.go.id

Rozeha A. Rashid
*Telecommunication Software and Systems Research Group,*
*Faculty of Electrical Engineering,*
*Universiti Teknologi Malaysia (UTM),*
Johor Bahru, Johor, Malaysia
rozeha@utm.my

*Abstract*—This paper describes the implementation of a learning-based lane detection algorithm on an autonomous mobile robot. More specifically, it explores the implementation of the Ultra Fast Lane Detection algorithm for real-time application on the SEATER (Single-passenger Electric Autonomous Transporter) prototype using a camera, with an emphasis on optimizing its performance on a Jetson Nano platform. Multi-core processing on the CPU and GPU has been implemented in this paper to optimize the application program and avoid delays or buffering. Preliminary experiments have been conducted to evaluate the algorithm's performance in terms of data processing speed and accuracy using two datasets: a public dataset for the outdoor environment and another dataset collected from the indoor area of the Workshop Building at BRIN Bandung. The experimental results show that after conversion to TensorRT, the algorithm achieves significantly improved performance on the Jetson Nano platform compared to the ONNX model, with processing speeds of approximately 101 ms using CULane and 105 ms using TuSimple—about 22 times faster than the previous model. While the algorithm demonstrates good accuracy on the public outdoor dataset, its performance falls short on the indoor dataset. Future work should focus on transfer learning and fine-tuning to improve indoor lane detection accuracy.

I. INTRODUCTION

Autonomous mobile robots are transforming how we navigate and move through different spaces, making it possible to travel comfortably in a vehicle without continuous human intervention. Research in autonomous mobile robots has been rapidly expanding in recent years, with one notable innovation is the SEATER (Single-passenger Electric Autonomous Transporter) platform, designed to enhance mobility for individuals by providing a single-passenger transport solution [1]. SEATER is a wheelchair-shaped mobile robot that can accommodate a single passenger and move either manually using a joystick or autonomously within specific environments. The SEATER's navigation system is built on cutting-edge technologies such as simultaneous localization and mapping (SLAM) and the Robot Operating System (ROS), ensuring precise and reliable operation [2]. To enhance SEATER's navigation system and improve its accuracy, lane detection can be a valuable method to added to the system. Lane detection is a technique for identifying the lane markings that guide a mobile robot's path, ensuring it stays within the designated lane. Additionally, the lane detection results can aid in localization and correcting the vehicle's position and state estimation [3]. Numerous research papers have explored lane detection methods, including both non-learning-based approaches [4][5] and learning-based methods, such as image segmentation, which processes camera-captured images. LaneNet [6] stands as a pioneer in this field, followed by the emergence of other methods like SCNN [7], RESA [8], CurveLane-NAS [9], and Ultra Fast Lane Detection [10].

This paper aims to implement the Ultra Fast Lane Detection method for real-time lane detection on a low-power consumption data processing device. This method was chosen for its superior accuracy and processing speed, reportedly up to four times faster than previous state-of-the-art methods [10], where the original implementation of this method used PyTorch on an NVIDIA GTX 1080Ti GPU. The detailed contribution of this paper can be summarized as follows:

- This paper attempts to implement the Ultra Fast Lane Detection method on a low-power processing device, Jetson Nano, by converting the original model using TensorFlow into TensorRT and adding multicore processing to separate the image reading process, making the processing faster and avoiding buffering or delay.
- The datasets were collected for experimenting in two scenarios: (i) an outdoor scenario using public data and (ii) an indoor scenario using our dataset collected from the Workshop Building in BRIN Bandung.
- In addition, preliminary experiments have been conducted using pre-trained TuSimple and CULane models to evaluate the performance in terms of processing speed after conversion and the addition of multicore processing, as well as the accuracy of the model on the outdoor and indoor datasets.

The rest of this paper is organized as follows: Section 2 details the methodologies employed in the study. Section 3 presents the experimental results and offers a thorough analysis of the findings. Finally, Section 4 provides the conclusion drawn from the research and suggests potential directions for future work.

## II. METHODS

### A. SEATER Platform

In this study, we utilized the SEATER prototype as the research platform to explore and enhance autonomous navigation technology, with a particular focus on lane detection. SEATER, a wheelchair-shaped mobile robot, is designed as an autonomous vehicle that can provide mobility on-demand services and accommodate a single passenger. Fig.1 illustrates the general system diagram of SEATER as an autonomous mobility on-demand service.

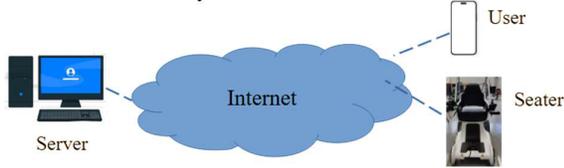

Fig. 1. System diagram of SEATER as autonomous mobility on-demand service.

As depicted in Fig. 1, the system is connected to the internet, allowing users to book and monitor the SEATER's status via their smartphones using the SEATER app installed. Once a user books the SEATER and selects their starting station, the SEATER will autonomously navigate to the user's starting station to pick up the passenger. The user can then choose the destination station from the user interface on the SEATER platform, and the SEATER will transport them to the selected destination. The SEATER can travel to multiple registered points or stations within a designated environment. Fig. 2 illustrates the SEATER prototype, which consists of two driving wheels connected to drive motors, two supporting omnidirectional wheels, a circuit, a data processing unit, a data communication media, and several sensors including a camera, which serves as the primary sensor for the autonomous navigation system. The primary objective of this paper is to conduct preliminary experiments evaluating the performance of the Ultra Fast Lane Detection method on the SEATER prototype for lane detection using data captured from a camera and processed on a Jetson Nano. This lane detection system will be integrated into the SEATER navigation system to ensure that the SEATER remain within its designated lane. The camera used for lane detection is a RealSense camera, as indicated by the circle in Fig. 2.

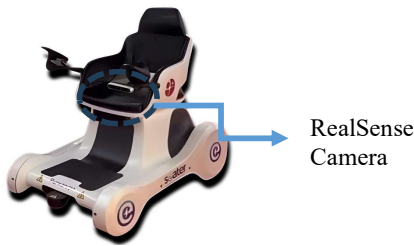

Fig. 2. SEATER Prototype

### B. Lane Detection Method

In the Ultra Fast Lane Detection method, lane detection is formulated as a row-based selection problem utilizing global features. The backbone of this method consists of lightweight versions of ResNet-18 and ResNet-34. Fig 3. Illustrates the architecture of the pre-trained model used in this work. The architecture processes the input image to produce an output image with detected lane positions. This model was originally developed using PyTorch and trained on the TuSimple and CULane datasets. TuSimple and CULane are large public datasets containing road images for lane detection. The TuSimple dataset consists of 6,408 road images from US highways, while CULane includes more than 55 hours of videos or 133,235 frames extracted from road images in Beijing [11][12]. In this study, we converted an ONNX (Open Neural Network Exchange) model into a TensorRT model for evaluation on a Jetson Nano. Fig. 4 illustrates the model conversion process used in the experiments in this work.

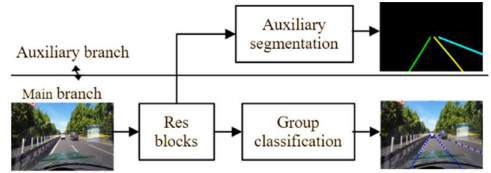

Fig.3. Overall architecture of Ultra Fast Lane Detection [10]

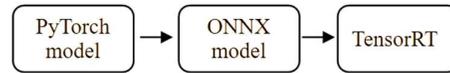

Fig. 4. Model conversion for experimenting on Jetson Nano

ONNX is an open-source artificial intelligence ecosystem that facilitates interoperability between various frameworks, tools, compilers, and runtimes. Meanwhile, TensorRT is a Software Development Kit (SDK) developed by NVIDIA and designed specifically for high-performance deep-learning inference [13][14]. Converting the model into TensorRT allows the Jetson Nano to fully utilize its GPU resources, thereby accelerating the execution of the application program. By converting the model into TensorRT, we aim to optimize the lane detection process on the SEATER prototype, ensuring that it can perform real-time lane detection.

For the experiments, two types of datasets are employed: outdoor and indoor scenarios. For the outdoor, we utilized a highway video dataset obtained from [15], while for the indoor, the dataset was collected using the SEATER's camera in the indoor area of the BRIN Workshop Building in Bandung. The outdoor dataset is used to validate the model's accuracy, as the pretraining model was trained using outdoor road images. To ensure robust evaluation, a new dataset, different from the commonly used TuSimple and CULane datasets is used for outdoor scenarios. Concurrently, an indoor dataset from the BRIN Workshop Building in Bandung is used as this is the intended operational environment for the SEATER prototype. This comprehensive dataset consists of several videos from both outdoor and indoor scenarios, with the number of images extracted of over 100 images for outdoor scenes and more than 800 images for indoor scenes.

The flowchart for the lane detection experiment is depicted in Fig.5. The lane detection process begins with loading an image. A pre-trained model that has been loaded beforehand is then used to perform the lane detection on the image. The detected lanes are then superimposed onto the image and displayed on the screen. Based on the lane detection results, the system determines the position of the mobile robot relative to the lane. This information enables the system to issue commands to the drive motors, ensuring the robot moves according to the detected lane. For instance, if the mobile robot is positioned to the right of the lane, the system can instruct the drive motors to turn left, and vice versa, to keep the robot aligned within the lane. In addition, to accelerate the execution process and avoid overlap between processes, multithreading/multicore processing using the CPU is also employed in this paper, along with the GPU. As is known, the Jetson Nano has specifications including a 128-core Maxwell™ GPU and a quad-core ARM® Cortex®-A57 CPU. In this paper, the GPU is used to execute the deep learning model for lane detection, while other processes are run on the CPU. The detailed usage of Jetson Nano resources in this paper can be seen in Fig. 6.

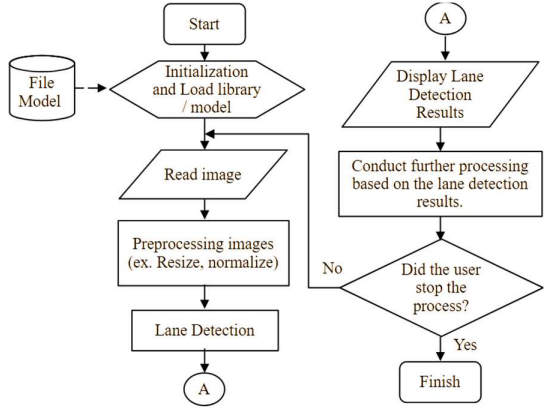

Fig. 5. Flowchart of the lane detection experiment

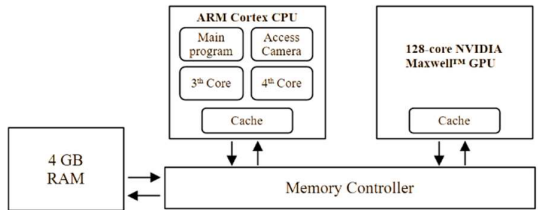

Fig. 6. Optimizing resource on Jetson Nano

## III. RESULTS AND DISCUSSION

A Python-based implementation code has been developed based on the methodologies presented in the references [16] [17]. The implementation includes three primary programs: the lane detection application, a video-to-image conversion tool, and an ONNX-to-TensorRT model conversion utility. The lane detection application is designed for real-time lane detection experiments, leveraging a pre-trained model to analyze images either captured directly from the SEATER's camera or loaded from files. Additionally, the video-to-image conversion tool facilitates dataset creation by extracting individual frames from video files, ensuring a comprehensive collection of images for training and testing. The ONNX-to-TensorRT model conversion utility optimizes the pre-trained models for deployment on the Jetson Nano, significantly enhancing inference speed and efficiency. Fig. 7 demonstrates the lane detection application interface developed in this work. The outdoor dataset, which consists of several highway videos which obtained from [15], and the indoor dataset, comprising videos captured in the indoor area of the BRIN Workshop Building in Bandung, were subsequently converted into images for use as testing datasets. Fig. 8 shows examples of the datasets employed in the experiments, highlighting the variety of images used to test the system's robustness across different environments. Table 1 shows the processing speeds observed during lane detection experiments conducted using both the ONNX model and the TensorRT-converted model on the test datasets. These experiments were carried out to compare the performance of each model.

Based on Table 1, it can be seen that the processing speed of the TensorRT model is significantly faster compared to the ONNX model. For instance, the TensorRT model processes TuSimple in 105 ms compared to 2,295 ms for the ONNX model, making it 21.86 times faster. Similarly, the TensorRT model processes CULane in approximately 101 ms, whereas the ONNX model takes 2,280 ms, which is 22.57 times faster. These results indicate that the TensorRT model has the potential to be implemented on SEATER using a Jetson Nano processing device.

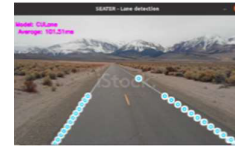

Fig. 7. Example of the interface of the lane detection application using CuLane which was developed in this work.

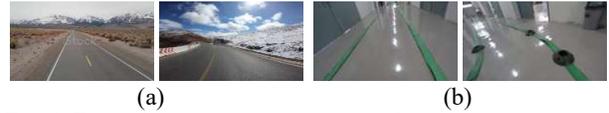

Fig. 8. Example of the dataset used: (a) Outdoor/highway; and (b) Indoor.

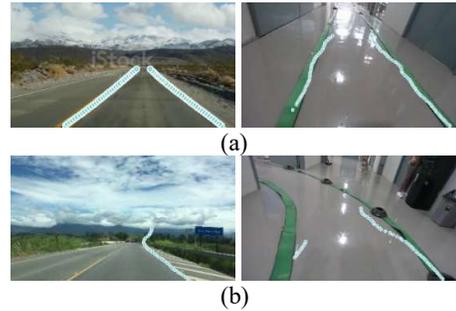

Fig. 9. Examples of lane detection results using TuSimple: (a) Detected lanes; (b) Undetected lanes

TABLE I. DATA PROCESSING SPEED BEFORE AND AFTER CONVERSION

|  | ONNX model | | TensorRT model | |
|---|---|---|---|---|
|  | *TuSimple* | *CULane* | *TuSimple* | *CULane* |
| Data Processing Speed (ms) | 2,295 | 2,280 | 105 | 101 |

TABLE II. PRETRAINING MODEL ACCURATION

| | Outdoor/highway data | | | | Indoor data | | | |
|---|---|---|---|---|---|---|---|---|
| | *TuSimple* | | *CULane* | | *TuSimple* | | *CULane* | |
| Lane detected or Undetected (√ / -) | √ | √ | √ | √ | - | - | √ | - |
| | √ | √ | √ | √ | - | - | - | √ |
| | √ | √ | - | √ | √ | √ | √ | - |
| | - | √ | - | √ | √ | √ | - | - |
| | √ | √ | - | √ | √ | - | - | - |
| | √ | √ | √ | √ | - | - | √ | - |
| | √ | √ | √ | √ | - | - | - | - |

The experiment also revealed buffering or delay when reading images from the camera. This issue arises because the lane detection inference process is not as fast as the camera's image-capturing process, resulting in the accessed image being not the most recent but an image from a previous time. To tackle this issue, the utilization of multicore processing using the CPU is employed to separate the image reading process from the camera onto different cores as seen in Fig. 6. Additionally, adjusting the frame rate of image acquisition from the camera to match the lane detection processing time (approximately 9 fps) or setting the buffer size (based on experiments, the most optimal value is 1) can also be implemented to address this issue. After evaluating the processing speed, an additional experiment was conducted to assess the accuracy of the pre-trained model. The results of this accuracy evaluation are presented in Table 2, providing insights into the model's performance in accurately detecting lanes across different datasets.

The criteria for categorizing the lane detection results as either "detected" or "undetected" in this paper are as follows: if both the left and right lanes are detected, even if imperfectly, but their direction is not deviating, they are considered detected. If this condition is not met, the result is considered undetected. Based on Table 2, it can be found that the method performs well on the outdoor dataset, demonstrating high accuracy in lane detection. However, the results for the indoor dataset are significantly less accurate. Examples of lane detection results using TuSimple are shown in Fig. 9 whereas results using CULane can be found in Fig. 7. Based on experiments, it can be concluded that for indoor lane detection, while some lanes are detected, none of the results perfectly detect the left and right lanes as seen in the right side of Fig. 9.a. This discrepancy is likely due to the training dataset used for this method, which consists of outdoor highway datasets, whereas the indoor dataset used in the experiment features different lane shapes and contains noise such as light reflections and lane boundary markers. To enhance the method's accuracy, transfer learning and fine-tuning using an indoor dataset could be conducted in future work.

## IV. Conclusion

This paper presents the implementation and evaluation of the Ultra Fast Lane Detection algorithm on an Autonomous Mobile Robot. By optimizing the algorithm's performance on a Jetson Nano platform, we aimed to achieve real-time lane detection using a camera. Our experiments utilized two types of datasets: a public dataset for outdoor environments and an internally collected dataset from the indoor area of the Workshop Building in BRIN Bandung. The experimental results demonstrate that converting the model to the TensorRT model significantly enhanced its processing speed on the Jetson Nano platform, achieving processing times of approximately 101 ms with CULane and 105 ms with TuSimple—an improvement of about 22 times compared to the ONNX model. This superior performance is attributed to the TensorRT model's ability to leverage the Jetson Nano's GPU resources for faster processing. Additionally, the issue of delay or buffering when reading images from the camera can also be addressed by utilizing multicore processing on the CPU. While the algorithm performed well in terms of accuracy on the outdoor public dataset, its performance on the indoor dataset was less satisfactory. To address this discrepancy and improve indoor lane detection accuracy, future work should focus on employing transfer learning and fine-tuning techniques to enhance the model's robustness and reliability, enabling effective real-time lane detection in varied environments.

## Acknowledgment

This work is supported by a research project funded by the Research Organization for Electronics and Informatics, National Research and Innovation Agency. The authors would like to thank to the Faculty of Electrical Engineering, Universiti Teknologi Malaysia (UTM) and the Research Center for Smart Mechatronics, National Research and Innovation Agency (BRIN), as well as everyone who assisted in this research.

## References

[1] R. P. Saputra, M. Mirdanies, E. J. Pristianto, and D. Kurniawan, "Autonomous Docking Method via Non-linear Model Predictive Control," *Proceeding - 2023 Int. Conf. Radar, Antenna, Microwave, Electron. Telecommun. Empower. Glob. Prog. Innov. Electron. Telecommun. Solut. a Sustain. Futur. ICRAMET 2023*, pp. 331–336, 2023, doi: 10.1109/ICRAMET60171.2023.10366563.

[2] A. S. Irwansyah, B. Heryadi, D. K. Dewi, R. P. Saputra, and Z. Abidin, "ROS-based multi-sensor integrated localization system for cost-effective and accurate indoor navigation system," *Int. J. Intell. Robot. Appl.*, pp. 1–19, Jun. 2024, doi: 10.1007/S41315-024-00350-1/TABLES/2.

[3] R. Sadli, M. Afkir, A. Hadid, A. Rivenq, and A. Taleb-Ahmed, "Map-Matching-Based Localization Using Camera and Low-Cost GPS For Lane-Level Accuracy," *Procedia Comput. Sci.*, vol. 198, pp. 255–262, Jan. 2022, doi: 10.1016/J.PROCS.2021.12.237.

[4] S. Kumar, M. Jailia, and S. Varshney, "An efficient approach for highway lane detection based on the Hough transform and Kalman filter," *Innov. Infrastruct. Solut. 2022 75*, vol. 7, no. 5, pp. 1–24, Jul. 2022, doi: 10.1007/S41062-022-00887-9.

[5] Q. Huang and J. Liu, "Practical limitations of lane detection algorithm based on Hough transform in challenging scenarios," *Int. J. Adv. Robot. Syst.*, vol. 18, no. 2, Apr. 2021, doi: 10.1177/17298814211008752/ASSET/IMAGES/LARGE/10.1177_17298814211008752-FIG11.JPEG.

[6] D. Neven, B. De Brabandere, S. Georgoulis, M. Proesmans, and L. Van Gool, "Towards End-to-End Lane Detection: An Instance Segmentation Approach," *IEEE Intell. Veh. Symp. Proc.*, vol. 2018-June, pp. 286–291, Oct. 2018, doi: 10.1109/IVS.2018.8500547.

[7] X. Pan, J. Shi, P. Luo, X. Wang, and X. Tang, "Spatial as Deep: Spatial CNN for Traffic Scene Understanding," doi: 10.5555/3504035.3504926.

[8] T. Zheng *et al.*, "RESA: Recurrent Feature-Shift Aggregator for Lane Detection," *Proc. AAAI Conf. Artif. Intell.*, vol. 35, no. 4, pp. 3547–3554, May 2021, doi: 10.1609/AAAI.V35I4.16469.

[9] H. Xu, S. Wang, X. Cai, W. Zhang, X. Liang, and Z. Li, "CurveLane-NAS: Unifying Lane-Sensitive Architecture Search and Adaptive Point Blending," *Lect. Notes Comput. Sci. (including Subser. Lect. Notes Artif. Intell. Lect. Notes Bioinformatics)*, vol. 12360 LNCS, pp. 689–704, 2020, doi: 10.1007/978-3-030-58555-6_41.

[10] Z. Qin, H. Wang, and X. Li, "Ultra Fast Structure-Aware Deep Lane Detection," *Lect. Notes Comput. Sci. (including Subser. Lect. Notes Artif. Intell. Lect. Notes Bioinformatics)*, vol. 12369 LNCS, pp. 276–291, 2020, doi: 10.1007/978-3-030-58586-0_17.

[11] "GitHub - TuSimple/tusimple-benchmark: Download Datasets and Ground Truths: https://github.com/TuSimple/tusimple-benchmark/issues/3." https://github.com/TuSimple/tusimple-benchmark (accessed Aug. 19, 2024).

[12] "CULane Dataset." https://xingangpan.github.io/projects/CULane.html (accessed Aug. 19, 2024).

[13] Y. Zhou, Z. Guo, Z. Dong, and K. Yang, "TensorRT Implementations of Model Quantization on Edge SoC," *Proc. - 2023 16th IEEE Int. Symp. Embed. Multicore/Many-Core Syst. MCSoC 2023*, pp. 486–493, 2023, doi: 10.1109/MCSOC60832.2023.00078.

[14] "TensorRT SDK | NVIDIA Developer." https://developer.nvidia.com/tensorrt (accessed Jun. 21, 2024).

[15] "6.000+ Video, Klip HD & 4K Jalan Raya & Jalan Gratis - Pixabay." https://pixabay.com/id/videos/search/jalan raya/ (accessed Jun. 21, 2024).

[16] "GitHub - MaybeShewill-CV/lanenet-lane-detection: Unofficial implementation of lanenet model for real time lane detection." https://github.com/MaybeShewill-CV/lanenet-lane-detection?tab=readme-ov-file (accessed Jun. 21, 2024).

[17] "GitHub - ibaiGorordo/onnx-Ultra-Fast-Lane-Detection-Inference: Example scripts for the detection of lanes using the ultra fast lane detection model in ONNX." https://github.com/ibaiGorordo/onnx-Ultra-Fast-Lane-Detection-Inference (accessed Jun. 21, 2024).